\RequirePackage{fix-cm}
\documentclass[twocolumn,envcountsect]{svjour3}          
\smartqed  
\usepackage{graphicx}
\usepackage{amsmath}
\usepackage{amssymb}
\usepackage{booktabs}
\usepackage{multirow}
\usepackage{caption}
\usepackage{algorithm}
\usepackage{algorithmic}
\usepackage{adjustbox}
\usepackage{blindtext}
\usepackage{rotating}
\usepackage{threeparttable}
\usepackage[pagebackref,breaklinks,colorlinks]{hyperref}
\usepackage{hyperref}
\usepackage[capitalize]{cleveref}
\hypersetup{hidelinks,
	colorlinks=true,
	allcolors=black,
	pdfstartview=Fit,
	breaklinks=true}
\crefname{section}{Sec.}{Secs.}
\Crefname{section}{Section}{Sections}
\Crefname{table}{Table}{Tables}
\crefname{table}{Tab.}{Tabs.}

\begin{document}
\begin{sloppypar}

\title{Image Understands Point Cloud: Weakly Supervised 3D Semantic Segmentation via Association Learning}


\author{Tianfang Sun \and
        Zhizhong Zhang \and
        Xin Tan \and
        Yanyun Qu \and
        Yuan Xie \and
        Lizhuang Ma
}


\institute{Tianfang Sun \at
           East China Normal University, Shanghai, China \\
           \email{52205901023@stu.ecnu.edu.cn} \and
           Zhizhong Zhang \at
           East China Normal University, Shanghai, China \\
           \email{zzzhang@cs.ecnu.edu.cn} \and
           Xin Tan \at
           East China Normal University, Shanghai, China \\
           \email{xtan@cs.ecnu.edu.cn} \and
           Yanyun Qu \at
           Xiamen University, Xiamen, Fujian, China \\
           \email{yyqu@xmu.edu.cn} \and
           Yuan Xie \at
           East China Normal University, Shanghai, China \\
           \email{xieyuan8589@foxmail.com} \and
           Lizhuang Ma \at
           Shanghai Jiao Tong University, Shanghai, China \\
           \email{ma-lz@cs.sjtu.edu.cn}
}

\date{Received: date / Accepted: date}

\maketitle

\begin{abstract}
Weakly supervised point cloud semantic segmentation methods that require 1\% or fewer labels, hoping to realize almost the same performance as fully supervised approaches, which recently, have attracted extensive research attention. A typical solution in this framework is to use self-training or pseudo labeling to mine the supervision from the point cloud itself, 
but ignore the critical information from images. In fact, cameras widely exist in LiDAR scenarios and this complementary information seems to be greatly important for 3D applications. 
In this paper, we propose a novel cross-modality weakly supervised method for 3D segmentation, incorporating complementary information from unlabeled images. Basically, we design a dual-branch network equipped with an active labeling strategy, to maximize the power of tiny parts of labels and directly realize 2D-to-3D knowledge transfer. Afterwards, we establish a cross-modal self-training framework in an Expectation-Maximum (EM) perspective, which iterates between pseudo labels estimation and parameters updating. In the M-Step, we propose a cross-modal association learning to mine complementary supervision from images by reinforcing the cycle-consistency between 3D points and 2D superpixels. In the E-step, a pseudo label self-rectification mechanism is derived to filter noise labels thus providing more accurate labels for the networks to get fully trained. The extensive experimental results demonstrate that our method even outperforms the state-of-the-art fully supervised competitors with less than 1\% actively selected annotations.
\keywords{multi-modal \and weakly supervised \and point cloud semantic segmentation}
\end{abstract}

\begin{figure}[t]
  \centering
  \includegraphics[width=0.99\linewidth]{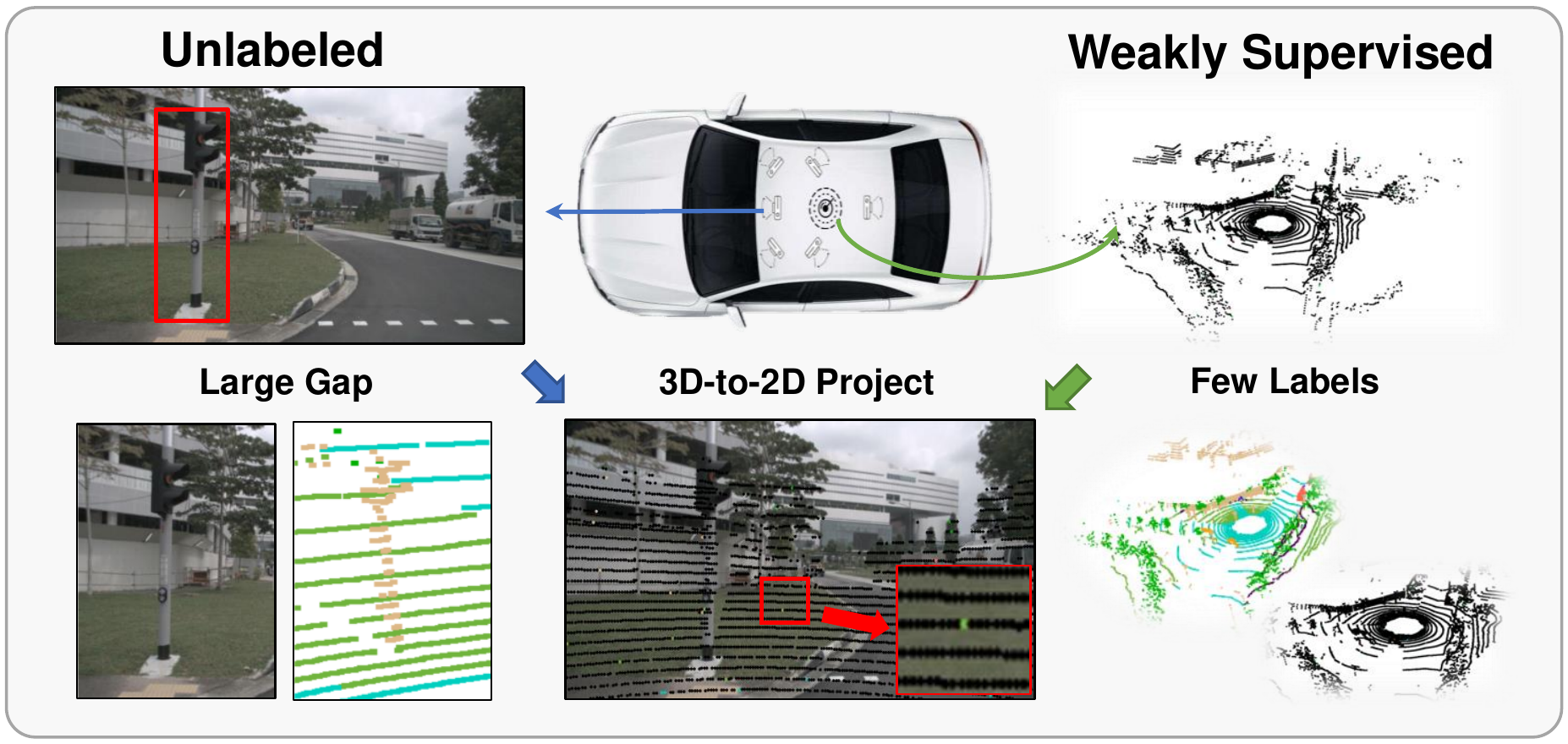}
  \caption{The main challenges exist in cross-modal weakly supervised learning. The modality gaps between LiDAR point clouds and images are large but their annotated labels are few.} 
  \label{fig:intro}
\end{figure}

\section{Introduction}\label{sec:intro}

Point cloud provides a powerful way to perceive, understand and reconstruct the complex 3D visual world. It reflects the fine-grained environmental information for detecting and recognizing objects \cite{qi2017pointnet,qi2017pointnet++,hu2020randla} with accurate depth information. Lying in the heart of point cloud vision, semantic segmentation plays a critical role in automatic driving, particularly in robotics and virtual reality \cite{gan2019self,shan2018lego,liu2019lpd}.

Recently, fully supervised methods \cite{tang2020searching,cortinhal2020salsanext,zhuang2021perception} in this community have achieved very promising performance. One of the ingredients to their success was the availability of well-annotated training sets. However, it is exhausting to annotate point cloud data, which prohibitively restricts its potential applications. For example, on SemanticKITTI  \cite{behley2019semantickitti}, it takes on average 1.5-4.5 hours to label a single tile of point cloud, while it includes approximately 23,201 scenes in total, which is unacceptable in real-world applications. Instead, there has been an increasing interest in weakly supervised semantic segmentation \footnote{labeling a tiny fraction of points} \cite{zhang2021weakly,zhang2021perturbed,hu2021sqn}. 

An appealing way to this weakly supervised problem is the self-training method depending either on contrastive learning with data augmentation \cite{zhang2021semi,tarvainen2017mean,sohn2020fixmatch} or on additional data pretraining \cite{wang2021self,you2020co}. But their performances rely heavily on the auxiliary supervision mined from heuristic designs. Alternatively, utilizing the tiny labeled data to generate pseudo labels is another popular solution, but this kind of approach usually suffers from severe noise labels. By reviewing recent weakly supervised methods, we find an important issue which 
is not well addressed so far, {\it i.e.,} should we only consider mining the supervision from the point cloud data itself?

To tackle this issue, we have noticed that in a typical point cloud scenario, {\it e.g.,} automatic driving, there usually exist camera sensors that assist LiDAR sensor (see the top of \cref{fig:intro}). In this case, some images are associated with the LiDAR scene, which provides rich color and texture information. Furthermore, we can explicitly establish a point-to-pixel transformation connecting 3D and 2D data by synchronizing and calibrating between LiDAR and camera sensors. On the one hand, we can use the image features to enhance the point cloud representations yielding performance gain without any additional annotation cost. On the other hand, we are able to mine the complementary supervision from the data connection. 
 
Motivated by this, we propose a new yet practical setting for image understanding point cloud. In this setting, as shown in \cref{fig:intro}, we assume the point cloud is given or actively annotated with very few labels per category (1\% or 0.1\%), and the corresponding images are not labeled. With geometry transformation, the sparse labels from 3D points can be mapped to the corresponding 2D pixels, which enables us to explore their particular data merits and learn extra complementary supervision. 
Fortunately, the recently built multi-modality datasets \cite{caesar2020nuscenes,behley2019semantickitti,sun2020scalability} facilitate us to fuse 2D images and 3D point clouds with a positive learning effect.


 
However, there exist two principal problems in this setting. 1) \textit{Large modality gap but few labels.} Few labels magnify the modality gap (see the left bottom of \cref{fig:intro}, traffic lights), which makes the networks hard to optimize. 2) \textit{Imbalanced modality capability.} Annotations in pixels are much sparser than the points (see red box in the middle bottom of \cref{fig:intro}) which causes the performance of the 2D branch to be much weaker than the 3D branch. Therefore, a naive combination of the two kinds of data would degrade the performance.
 
In this paper, motivated by the automated labeling strategy \cite{cai2021revisiting,wu2021redal,liu2022less}, we provide an efficient yet effective labeling approach, which actively annotates a tiny part of labels to maximize the power of weak supervision. 
On this basis, we propose a cross-modal 3D segmentation dual-branch network, incorporating complementary information from unlabeled images. This network includes an image branch and a point cloud branch, which allows us to transfer knowledge directly by explicitly connecting 2D and 3D data by a perspective transformation. 
 
To further mine the complementary supervision for the 3D network from the much weaker 2D network, we establish a self-training framework from an Expectation-Maximum (EM) framework. In the M-Step, we maximize the log-likelihood function to update the network parameters. During it, we design an association learning method, taking advantage of supervision from image superpixels (pixels in visually similar regions usually contain similar semantic information). It eliminates the modality gap and helps the 3D branch to learn more discriminative features by encouraging the cycle-consistency between 3D points and their corresponding superpixels. With the mined cross-modal correlation, superpixel image prior is well learned and the complementary knowledge is hence explored, which, in fact, enables us to deal with the problem of ``large modality gap but few labels''.


In the E-step, we estimate the posterior probability to generate pseudo labels. In this step, we design an adaptive confidence thresholding module and a feature similarity filtering module to improve the accuracy of the estimated pseudo labels. As a result, the 2D weak branch is therefore boosted by learning the shared knowledge from the 3D branch via the EM framework, which helps the network to handle the ``imbalanced modality capability''. The experimental results demonstrate that our approach outperforms the state-of-the-art networks by utilizing images under the fully supervised setting and the performance is highly consistent with these methods when given less than 1\% annotations. Our contributions can be summarised as follows:
	\begin{itemize}
        \item We raise a new yet practical cross-modal weakly-supervised setting for point cloud semantic segmentation, where images are leveraged without additional annotations. In our approach, a dual-branch network with active labeling is proposed, which achieves state-of-the-art performance under both supervised and weakly supervised settings.
        \item We propose a new association learning module to take advantage of superpixel segmentation from images. It enables us to mine complementary knowledge from images regardless of the large modality gap and imbalanced modality capability, therefore enhancing the modality correlation and helping the network to learn more discriminative features.
        \item We introduce a self-training approach from the Expectation-Maximum (EM) framework. It alternatively performs parameter updating and label estimation. A pseudo label self-rectification approach is derived to filter out noise labels. Consequently, both two branches are boosted by learning the shared knowledge embedded in the reliable pseudo labels.
	\end{itemize}

\section{Related Work}
\textbf{Fully Supervised Uni-modal Learning.} This branch of work can mainly be divided into the camera-based method and the lidar-based method. For the camera-based semantic segmentation, FCN \cite{long2015fully} first designs an end-to-end fully convolutional architecture from the image classification network. Recently, significant improvements have been achieved by exploring multi-scale information \cite{chen2017deeplab,yu2021bisenet}, dilated convolution \cite{chen2017rethinking,mohan2021efficientps} and attention mechanisms \cite{huang2019ccnet,yuan2021ocnet}. However, camera-based methods are very sensitive to lighting conditions, therefore not robust in outdoor scenarios. LiDAR-based method can be divided into point-based method \cite{qi2017pointnet,qi2017pointnet++,hu2020randla}, projection-based method \cite{milioto2019rangenet++,cortinhal2020salsanext,zhang2020polarnet} and voxel-based method \cite{zhu2021cylindrical,tang2020searching,xu2021rpvnet}. PointNet \cite{qi2017pointnet} is the first point-based method, which proposed to extract features through multi-layer perception. RandLA-Net \cite{hu2020randla} modifies it to process large sparse outdoor LiDAR point clouds directly. However, point-based methods suffer performance degradation when the point cloud gets sparser and sparser. Projection-based methods explores effective projection method such as spherical projection \cite{milioto2019rangenet++,cortinhal2020salsanext}, polar projection \cite{zhang2020polarnet}, or both \cite{liong2020amvnet} to map the 3D point cloud into 2D images and leverage 2D convolution to extract features. Though much faster in processing speed, quantization error can not be avoided during the dimension reduction. Voxel-based methods convert the point cloud to 3D voxels and utilize 3D convolution to extract the feature. These methods are less sensitive to point density and introduce less quantization error. Thanks to sparse convolution, the computation and memory consumption are largely reduced. Therefore, voxel-based methods are both effective and efficient. In this paper, we choose SPVCNN \cite{tang2020searching} as our 3D backbone.

\textbf{Fully Supervised Cross-modal Learning.} To combine the merits of the two modalities, many researchers try to enhance the point cloud feature with the knowledge from images by fusing the 2D and their corresponding 3D features \cite{krispel2020fuseseg,el2019rgb,meyer2019sensor}. To find the correspondence between points and pixels, a typical solution is to directly project the points to the image with a predefined calibration matrix \cite{vora2020pointpainting,huang2020epnet}, or through a bird-eye-view (BVE) projection \cite{yoo20203d}. The features can be fused by directly connecting \cite{vora2020pointpainting} or other well-designed ways \cite{yoo20203d,huang2020epnet}. Recently, PMF \cite{zhuang2021perception} exploits a collaborative fusion of multimodal data in camera coordinates. Although noticeable performance gain could be brought by these methods, they always rely on the fully supervised setting, requiring densely annotated data to provide supervision for network training and even demand additional labels to train the 2D network. In contrast, our method is able to leverage the knowledge from the unlabeled images and can be easily applied to a variety of weakly supervised scenarios. 

\textbf{Weakly Supervised Learning.} Previous works have dealt with the weakly supervised point cloud semantic segmentation problem either by using a small part of labels per category/instance \cite{zhang2021weakly,zhang2021perturbed,liu2021one} or mining the consistency regulation within the point cloud itself \cite{xie2020pointcontrast,hou2021exploring}. For example, Zhang et al. \cite{zhang2021weakly} design a point cloud colorization pretext task to assist the weakly supervised learning process. In PSD \cite{zhang2021perturbed}, a self-supervised method is proposed to establish consistency between the original point cloud and the perturbed one. In \cite{liu2021one}, the sparse labels are propagated according to the feature similarity with a conditional random field in an iterative way. These aforementioned methods make good use of the knowledge from the point cloud but are only proven to be effective in indoor scenes. However, due to the large difference in the color availability, density, and quantity of points, these methods may not be suitable for outdoor LiDAR point clouds. Recently, SLidR \cite{sautier2022image} develops a superpixel-based cross-modal contrastive learning method for outdoor scenario. However, its image branch has to be frozen to avoid collapse during training thus the information from images is not fully used. However, in our method, the two branches can be trained synchronistically. Moreover, our method does not require carefully designed perturbation or graph-based label propagation, but it is able to handle the challenging sparse outdoor LiDAR point cloud scenes where we mine the consistency supervision from images.

\begin{figure*}[htbp]
\begin{center}
\includegraphics[width=0.95\textwidth]{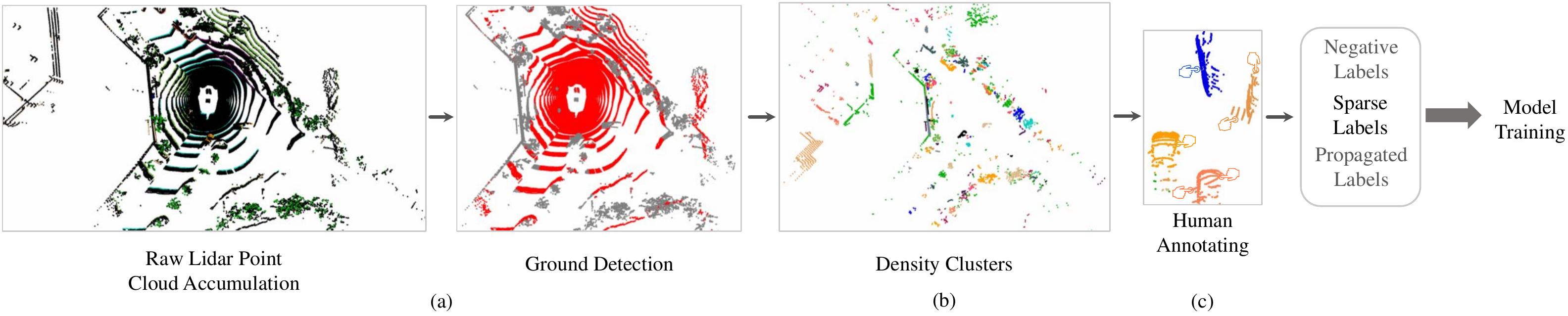}
\end{center}
\caption{Pipeline of active labeling. a) ground detection, b) density clustering and c) human annotation.}
\label{fig:pre_seg}
\end{figure*}

\section{Method}
Our goal is to develop a weakly supervised 3D segmentation method, borrowing complementary information from images. 
To this end, we design a simple yet effective point cloud segmentation baseline (Sec. \ref{sec:1}), where a self-training framework based on the EM algorithm is proposed (Sec. \ref{sec:2}). In the M-step, we introduce a cross-modal association method to connect 2D and 3D visual data and mine their supervision (Sec. \ref{sec:3}). In the E-step, we introduce a pseudo label self-rectification method to filter pseudo labels (Sec. \ref{sec:4}).

\begin{figure}[t]
  \centering
  \includegraphics[width=0.99\linewidth]{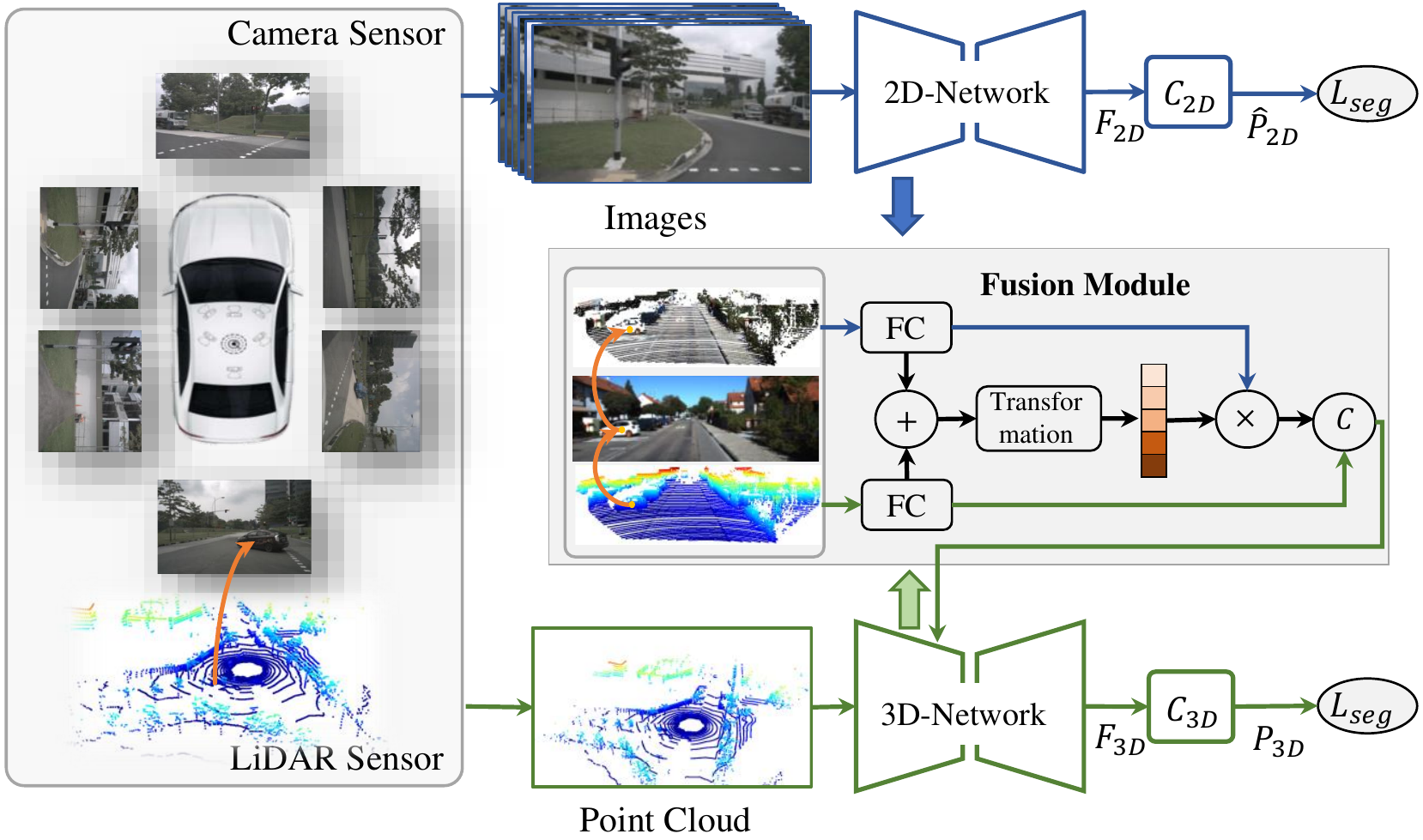}
  \caption{Cross-modal semantic segmentation baseline. Knowledge is directly transferred from the 2D network to the 3D network layer by layer through the feature fusion module.}
  \label{fig:baseline}
\end{figure}

\subsection{New Problem Setting and Baseline}\label{sec:1}
\textbf{Problem Setting.} We use $\boldsymbol{x}_{\text{3D}}$ 
to denote a 3D LiDAR point cloud, 
and each point in $\boldsymbol{x}_{\text{3D}}$ corresponds to a label $y \in [1,2,\ldots,C]$, where $C$ denotes the number of categories. Besides, in a typical point cloud scenario, {\it e.g.,} automatic driving, there usually exist cameras that assist LiDAR sensor to perceive the surrounding environment, such that images $\boldsymbol{x}_\text{2D}$ are provided and associated with the LiDAR scene (see \cref{fig:baseline}). For example, on the nuScenes \cite{caesar2020nuscenes} dataset, there are six images corresponding to a frame of the point cloud, so we can establish a perspective transformation between points and pixels, to further explore the complementary information contained in these data.


To avoid extensive annotation, we assume very few labels of point clouds are given ({\it e.g.,} 1\% or fewer), and each category has at least one labeled point. 2D images $\boldsymbol{x}_{\text{2D}}$ are without any annotations. To reduce the labeling efforts, we explore two labeling strategies including random labeling \cite{zhang2021perturbed}\cite{hu2021sqn}\cite{liu2021one}, which randomly annotates a tiny part of points, and active labeling, which is designed to annotate a tiny part of points selectively. 
Notice that this active labeling strategy is tightly linked to the real-world weakly supervised 3D applications, and our experimental results highlight its critical role, where we can achieve much better performance with fewer annotations. 

As illustrated in \cref{fig:pre_seg}, this active labeling contains the following three steps: \\
1) \textit{Ground Detection.} For each keyframe, given the provided ego-poses, we fuse consecutive scans to get the accumulated point cloud (see \cref{fig:pre_seg} (a)). Then, we split the point cloud into uniform pillars in the cylinder coordinates. Finally, we apply the RANSAC algorithm \cite{fischler1981random} to detect the ground points in each local cell. \\ 
2) \textit{Density Clustering.} We use a pre-segmentation approach HDBSCAN algorithm \cite{campello2013density} to accurately cluster points to instances after removing the detected ground points (see \cref{fig:pre_seg} (b)).\\
3) \textit{Human Annotation.} Finally, motivated by LESS \cite{liu2022less}, annotators only need to annotate each cluster coarsely. Take \cref{fig:pre_seg} (c) as an example, 
all the points in the blue cluster belong to one category (car). Therefore, only a single click is needed. However, points in the brown cluster belong to two categories, car and road. Thus one point label for each category is required. 

In summary, there exist three kinds of labels. \textbf{Sparse label} are the labels manually labeled by the annotators. \textbf{Propagated label} indicates the points in the clusters where only one category appears and we can propagate the label to the entire cluster. \textbf{Negative label} indicates the clusters including the points that belong to multiple categories which could help us to filter some incorrect labels. Different from LESS, we find the pillar division and HDBSCAN are easier to implement and good at handling a wide range of point densities.
The statistics of these kinds of labels are summarized in Table \ref{labelsts}.
\begin{table}[]
\caption{Statistics of active labeling. The rate of the sparse label is computed against the entire point cloud to show the label usage. The rates of the propagate label and negative label are computed against the whole training set.} \label{labelsts}
\resizebox{\linewidth}{!}{
\begin{tabular}{@{}c|ccc@{}}
\toprule
Statistics       & nuScenes & Waymo & SemanticKITTI \\ \midrule
Sparse label     & 0.8\%    & 0.3\% & 0.08\%         \\
Propagate label  & 48.5\%   & 21.2\%  & 22.7\%        \\
Negative label   & 44.5\%   & 76.7\%  & 70.6\%        \\ \bottomrule
\end{tabular}
}
\label{tab:al}
\end{table}

\textbf{Baseline Clarification.} We build a cross-modal 3D segmentation baseline to explicitly enhance the point cloud features with the corresponding image features. As illustrated in Fig. \ref{fig:baseline}, it contains the following components: \\
1) \textit{Individual Backbones.} Two U-net architectures named 2D network (SwiftNet \cite{orsic2019defense}) and 3D network (SPVCNN \cite{tang2020searching} ) are adopted to learn visual representations for image and LiDAR data, respectively. They are chosen due to the balance of performance and efficiency. Besides, we send $\boldsymbol{x}_{\text{3D}}$ and $\boldsymbol{x}_{\text{2D}}$ from the same scene into the networks to produce consistent representations. \\
2) \textit{Modality Fusion.} In the middle of the 2D and the 3D encoder, we adopt the LIFusion module proposed in EPNet \cite{huang2020epnet} to fuse two kinds of features from the corresponding layer. 
In our implementation,  for a point $\tilde{\boldsymbol{x}}_{3D}=(x, y, z, 1)^T$ with homogeneous coordinate, we compute the projected point $\tilde{\boldsymbol{x}}_{3D}=(\tilde{x}, \tilde{y}, \tilde{z})^T$ in pixel coordinates by the transformation matrix.  With this, we can find the corresponding image pixel and therefore obtain the 3D feature $\boldsymbol{F}_{3D}$ and 2D feature $\boldsymbol{F}_{2D}$. We fuse the features by
 \begin{equation}
     \begin{gathered}
         \boldsymbol{w}=\sigma(h(tanh(f(\boldsymbol{F}_{3D})+g(\boldsymbol{F}_{2D})))) \\
         \boldsymbol{F}_{fuse}=\boldsymbol{F}_{3D} \oplus \boldsymbol{wF}_{2D},
     \end{gathered}
 \end{equation}
 where $f$, $g$ and $h$ stand for multilayer perceptron (MLP) and $\oplus$ denotes the concatenation. As a result, each 3D feature is enhanced by mixing its aligned 2D feature. \\
3) \textit{Loss Function.} We optimize the network by cross-entropy and Lovasz-softmax loss \cite{berman2018lovasz} on labeled data. 
When applying active labeling, the learning objective consists of two parts. One is cross-entropy and Lovasz-softmax loss computed on sparse and propagated labeled data, and the other is computed on negative labeled data expressed as
\begin{equation}
    \mathcal{L}_{neg} = -\frac{1}{N}\sum\limits_{i=1}^{N}\log(1-\sum\limits_{c_{ij}=0}{p_{ij}}),
\end{equation}
where $p_{ij}$ is the prediction logit of point $i$ for category $j$. $c_{ij}=1$ denotes that the cluster including point $i$ contains category $j$. We refer to this process as supervised segmentation training, denoted as $\mathcal{L}_{seg}$.

\begin{figure*}[htbp]
\begin{center}
\includegraphics[width=0.95\textwidth]{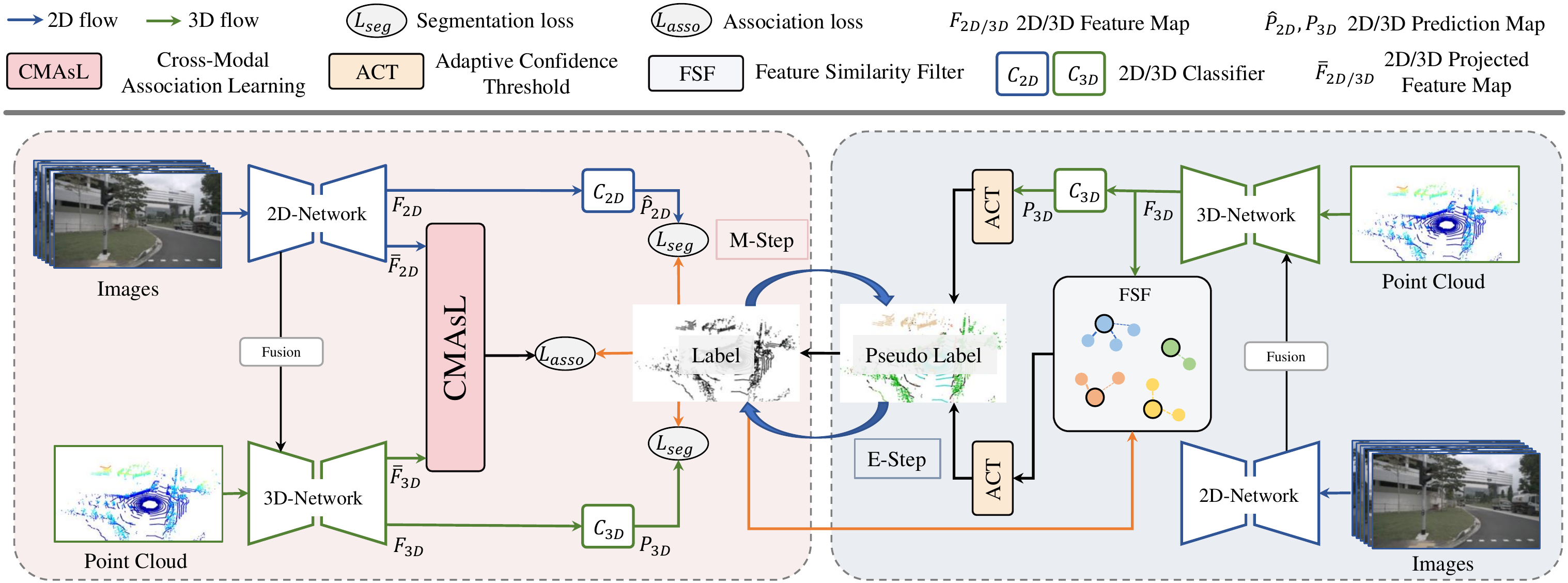}
\end{center}
\caption{The overall architecture of the proposed method. In the M-step (marked as pink), dimension-aligned features ($\bar{F}_{3D}$ and $\bar{F}_{2D}$) are fed into the cross-modal association learning module to reinforce the cycle-consistency between two modalities. In the E-step (marked as blue), pseudo labels are generated using the prediction of the 3D network ($P_{3D}$) and filtered with the adaptive confidence threshold and feature similarity filter.}
\label{fig:overall-architecture}
\end{figure*}

\subsection{Cross-modal Self-training Overview}\label{sec:2}
To train this cross-modality baseline, there exist two major challenges: 1) \textit{Large modality gap but few labels.} When given very few annotated data, both 2D and 3D branches are not fully trained. Feature fusion fails to mine the complementary supervision from two kinds of data. 2) \textit{imbalanced modality capability.} The annotations in the 2D branch are extremely sparse (the number of pixels is much larger than the number of points), which leads to the imbalance in the modality capability {\it i.e.,} 3D network is much stronger than 2D network.

To deal with them, we propose a cross-modal weakly-supervised 3D semantic segmentation framework as shown in \cref{fig:overall-architecture}. We formulate it from an EM perspective. Broadly speaking, in the M-step, our goal is to find the network parameters $\boldsymbol{\theta}$ that maximize the log-likelihood function:
\begin{equation}
    \boldsymbol{\theta}^* = \mathop{argmax}\limits_{\boldsymbol{\theta}}\label{eq:src_target}{\sum\limits_{i=1}^N{\log{\sum\limits_{c=1}^C{p(\boldsymbol{x}^{(i)},\bar{y}^{(c)}|\theta)}}}},
\end{equation}
where $\boldsymbol{x}^{(i)}$ denote the $i$-th sample in total of $N$. We omit the subscript 2D/3D for simplicity. And we assume that the observed samples are related to a latent variable: the unknown semantic label $\bar{\boldsymbol{y}}=\{\bar{y}^{(c)}\}_{c=1}^C$.
Finally, our optimize objective is redirected to:
\begin{equation}
    \boldsymbol{\theta}^*=\mathop{argmax}\limits_{\boldsymbol{\theta}}\sum\limits_{i=1}^N\sum\limits_{c=1}^C{p(\bar{y}^{(c)}|\boldsymbol{x}^{(i)},\boldsymbol{\theta})\log{p(\boldsymbol{x}^{(i)},\bar{y}^{(c)}|\boldsymbol{\theta})}}\label{eq:mstep}.
\end{equation}
Notice that this objective is obtained by Jensen's inequality shown in the appendix. Hence, when given latent variable $\bar{\boldsymbol{y}}$, we can optimize the parameters to maximize the log-likelihood function described in \cref{eq:mstep}. In the E-step, we estimate the posterior probability $p(\bar{y}^{(c)}|\boldsymbol{x}^{(i)},\boldsymbol{\theta}_{3D})$ to generate the pseudo labels from the predictions of the network. These pseudo labels are then filtered by our proposed adaptive confidence thresholding and feature similarity filtering.

The self-training process is operated by iteratively performing E-step and M-step.  In our EM framework, we design a cross-modal association learning approach to take advantage of image superpixel segmentation, which enables us to tackle the modality gap problem and hence mine the complementary knowledge from images and LiDAR points. This EM framework also boosts the weak 2D branch to substantially improve segmentation results. Next, we will describe our EM framework in detail.

\begin{figure}[t]
  \centering
  \includegraphics[width=0.95\linewidth]{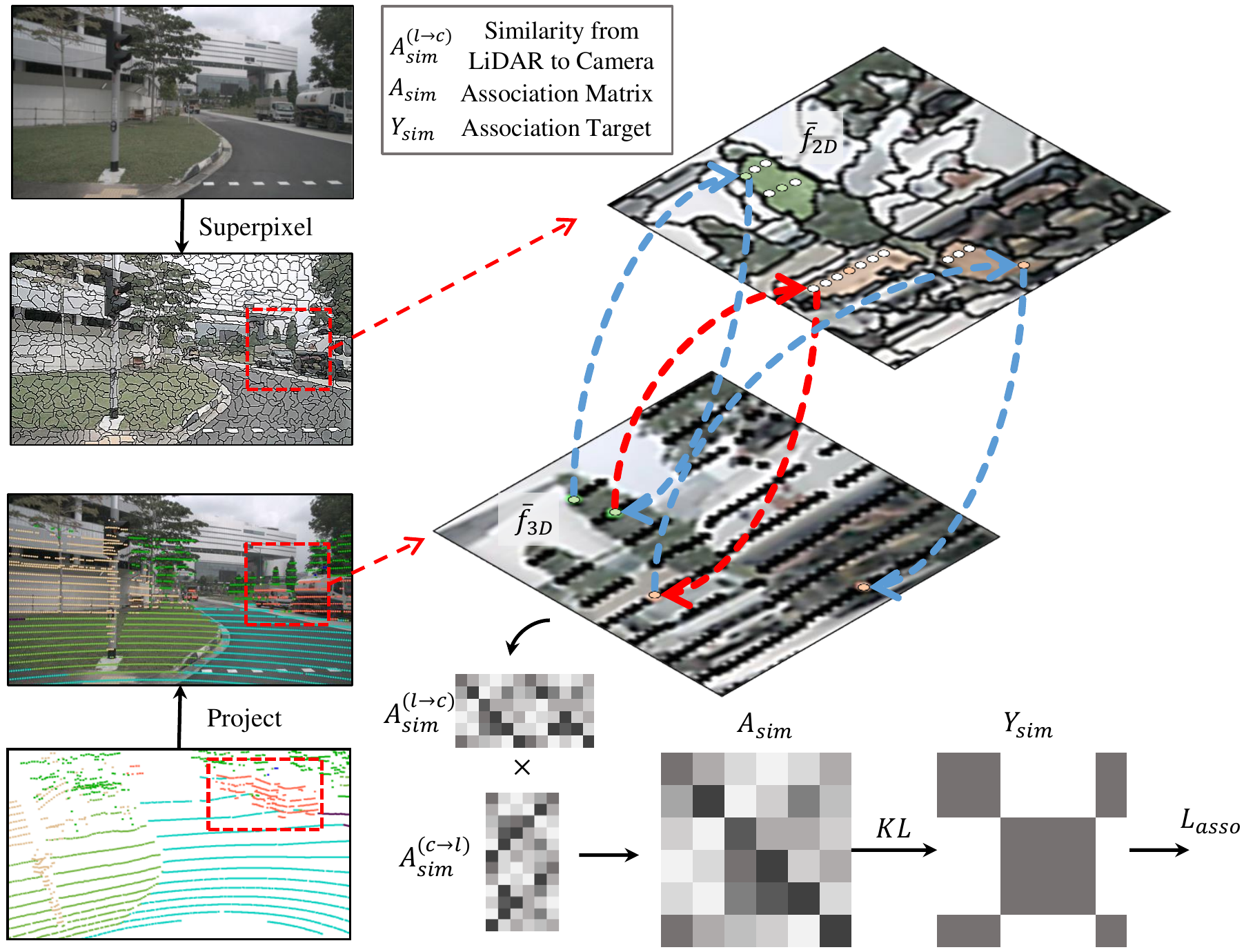}
  \caption{Illustration of the CMAsL module. The input images are firstly over-segmented into superpixels. The labeled 3D points are then projected to the image and find the superpixels that contain at least one point whose label is consistent with the 3D point. Transition probabilities are computed by calculating pairwise feature scalar product between 3D points and 2D points.}
  \label{fig:association}
\end{figure}

\subsection{M-step: Maximizing Likelihood via Cross-modality Association}\label{sec:3}
The goal of the M-step is to update the parameters with available data, but at the beginning of training, the amount of annotated data is very limited and the generated pseudo label is not accurate. In this case, the network tends to be overfitting and misleading. 
Therefore, to take full advantage of $\bar{\boldsymbol{y}}$, we propose (\textbf{C}ross-\textbf{M}odal \textbf{A}s\textbf{s}ociation \textbf{L}earning, \textbf{CMAsL}) module. The intuition is that we can explore the complementary knowledge from images, and its result in turn can facilitate the 3D segmentation.


As shown in \cref{fig:association}, we select SEEDS \cite{van2012seeds}, a training-free superpixel segmentation algorithm, to obtain a rough segmentation result ({\it i.e.,} superpixel). Each superpixel contains a small region of the image, and the pixels in this superpixel are highly locally consistent and probably belong to the same category. By taking this prior, CMAsL is designed to imagine a random walker traveling between 2D and 3D data. The walker starts from a labeled 3D point $\bar{\boldsymbol{f}}_{\text{3D}}$, visiting the 2D points $\bar{\boldsymbol{f}}_{\text{2D}}$ \footnote{we can map each 3D point to 2D coordinate, and the visited 2D data is from the same superpixel with at least one labeled 3D point.}, and ends at another labeled 3D point.
 
To set up an appropriate loss on this walk, we encourage 2D and 3D features with the same semantic information to be close to each other and hence promote the association between them. To this end, we first map the feature dimensions of 2D and 3D features to the same. Then we define a transition probability from $\bar{\boldsymbol{f}}_{\text{3D}}$ to $\bar{\boldsymbol{f}}_{\text{2D}}$: 
\begin{equation}
    a_{ij}^{(l\to c)}=\frac{\exp{(\langle\bar{\boldsymbol{f}}_{\text{3D}}^{(i)}, \bar{\boldsymbol{f}}_{\text{2D}}^{(j)}\rangle)}}{\sum_{j'}\exp{(\langle\bar{\boldsymbol{f}}}_{\text{3D}}^{(i)}, \bar{\boldsymbol{f}}_{\text{2D}}^{(j')}\rangle)},
\end{equation}
where $\langle \cdot, \cdot \rangle$ stands for the scalar product. We denote the transition matrix from 3D features to 2D features as $A^{(l\to c)}_{sim}\in \mathbb{R}^{N_l\times N_s}$ and the transition matrix $A^{(c\to l)}_{sim}\in \mathbb{R}^{N_s\times N_l}$ vice versa. $N_l$ and $N_s$ denote the number of 3D and 2D data. When the walker comes back, the associative similarity matrix $A_{sim}\in N_l\times N_l$ can be calculated as:
\begin{equation}
    A_{sim}=A^{(l\to c)}_{sim}\times A^{(c\to l)}_{sim}.
\end{equation}

We force the two-step probability to be similar to the uniform distribution over the class labels via a Kullback-Leibler Divergence:
\begin{equation}
    \mathcal{L}_{walker}=KL(Y_{sim}, A_{sim}),
\end{equation}
where
\begin{equation}
    Y_{sim}^{(ij)}=\left\{
    \begin{array}{cl}
    \frac{1}{|Y_{sim}^{(i)}|}     &  {y}_i={y}_j\\
    0     & else.
    \end{array}
    \right.
\end{equation}

Besides, to explore the hard samples, we additionally add a visit regulation $\mathcal{L}_{vis}$ to force all the points in $\bar{\boldsymbol{f}}_{\text{2D}}$ with equal probability to be visited. The association loss is finally expressed as:
\begin{equation}
    \mathcal{L}_{asso}=\beta_{w}\mathcal{L}_{walker}+\beta_{v}\mathcal{L}_{vis},
\end{equation}
where $\beta_{w},\beta_{v}$ are hyper-parameters. In our implementation, only 3D points with sparse labels are used as the start of association learning, as they are more representative both in spatial distribution and category distribution. Although propagate labels are almost as accurate as sparse labels, performing association with them is computationally prohibitive. Instead, to explore sample diversity, we perform association in the minibatch.
In summary, CMAsL is efficient, easy to implement, and model-agnostic.  
With the mined cross-modal correlation, superpixel image prior is well learned and the complementary knowledge is hence explored even when there exists a huge modality gap, without the need for additional supervision.


\subsection{E-step: Estimating Pseudo Label via Confidence Self-rectification}\label{sec:4}
In the E-step, we aim to generate reliable pseudo labels via a self-rectification mechanism and thereby helping the two branches to get better optimized. Although the network benefits from exploring the superpixel image prior, it still suffers from significant error aggregation raised by incorrect estimation. 
Therefore, we propose our (\textbf{A}daptive \textbf{C}onfidence \textbf{T}hreshold, \textbf{ACT}) module and (\textbf{F}eature \textbf{S}imilarity \textbf{F}ilter, \textbf{FSF}) module to improve the reliability of the pseudo labels. Note that as the sparse labels and the propagate labels are already accurate, we only generate pseudo labels from negative labeled points and unlabeled points.

\textbf{Adaptive Confidence Threshold.} The training dataset is highly class-imbalanced and the network tends to be confident in the dominant classes. Thus a fixed confidence threshold will either filter out too many accurate pseudo labels from minor classes or preserve many noisy pseudo labels from major classes. Therefore, we propose an adaptive confidence threshold to set different thresholds for different classes. Specifically, given the predicted class distribution $\boldsymbol{p} \in \mathbb{R}^{N\times C}$ and the raw pseudo one-hot labels $y_{\text{raw}}$, the confidence threshold for each category is calculated as:
\begin{equation}
    \sigma_c=\max(\max(\boldsymbol{p}_c)-\delta, \alpha), \forall c \in y_{\text{raw}},
\end{equation}
where $\boldsymbol{p}_c \in \mathbb{R}^N$ denotes a vector with elements representing c-$th$ category prediction for all $N$ samples. $\delta$ and $\alpha$ are two hyper-parameters standing for the confidence tolerance and confidence bottom line. With $\sigma_c$, a tighter threshold will be set to the categories that the network is confident with (\textit{e.g.} road, vegetation which enjoys more annotation points), and a looser threshold for hard categories (\textit{e.g.} pedestrian, bicycle which rarely have annotation points). As a result, ACT keeps a good balance between the accuracy and quantity of the pseudo labels.

\begin{algorithm}
    \renewcommand{\algorithmicrequire}{\textbf{Input:}}
    \renewcommand{\algorithmicensure}{\textbf{Output:}}
    \caption{Overall Procedure}
    \label{alg:main}
    \begin{algorithmic}[1]
        \REQUIRE Model $\mathcal{M}$ \\
                 Sparse labeled dataset $\mathcal{D}^s$ \\
                 Propagate labeled dataset $\mathcal{D}^p$ \\
                 Negative labeled dataset $\mathcal{D}^n$ \\
                 Unlabeled dataset $\mathcal{D}^u$ \\
                 Superpixel $\mathcal{S}$
        \ENSURE Trained model $\mathcal{M}$
        \STATE compute $\mathcal{L}_{seg}$ on ($\mathcal{D}^s\cup\mathcal{D}^p\cup\mathcal{D}^n$)
        \STATE compute $\mathcal{L}_{asso}$ on $\mathcal{D}^s$ and the matched superpixel $\mathcal{S}'$
        \STATE update $\mathcal{M}$ with $\mathcal{L}_{seg}$ and $\mathcal{L}_{asso}$
        \STATE $iter \leftarrow 0$
        \REPEAT
        \IF{$iter > 0$}
            \STATE generate pseudo labels $\mathcal{D}'$ on ($\mathcal{D}^n\cup\mathcal{D}^u$) using $\mathcal{M}$
        \ELSE
            \STATE generate pseudo labels $\mathcal{D}'$ on $\mathcal{D}^n$ using $\mathcal{M}$
        \ENDIF
        \STATE compute $\mathcal{L}_{seg}$ on ($\mathcal{D}^s\cup\mathcal{D}^p\cup\mathcal{D}^n\cup\mathcal{D}'$)
        \STATE compute $\mathcal{L}_{asso}$ on $\mathcal{D}^s$ and the matched superpixel $\mathcal{S}'$
        \STATE update $\mathcal{M}$ with $\mathcal{L}_{seg}$ and $\mathcal{L}_{asso}$
        \STATE $iter \leftarrow iter+1$
        \UNTIL performance not improved
        \RETURN{$\mathcal{M}$}
    \end{algorithmic}
\end{algorithm}

\textbf{Feature Similarity Filter.}  It is natural to have the idea of reducing the misleading pseudo labels by using its nearest class prototype, as points from the same category are usually located closer in the embedding space. Therefore, when the predictions from the classifier and the nearest prototype conflict, the label of the points will be filtered out. To be specific, given class prototypes $\boldsymbol{P}_{c}$ (averaging the feature of points with the same label), the predicted prototype label $\bar{y}^{p}_{i}$ of point $\boldsymbol{x}_i$ is defined as:
\begin{equation}
	\bar{y}^{p}_{i}=\mathop{argmax}\limits_{y_i}\frac{\exp(\langle f(\boldsymbol{x}_i),\boldsymbol{P}_{y_i}\rangle)}{\sum\limits_{c}\exp(\langle f(\boldsymbol{x}_i),\boldsymbol{P}_{c}\rangle)},
\end{equation}
where $f(\boldsymbol{x}_i)$ is the embedding feature. Note that $\bar{y}^p_i$ also passes through our adaptive confidence threshold module to filter the noise label. Afterwards, when $\bar{y}^c_i \neq \bar{y}^p_i$, where $\bar{y}^c_i$ is the prediction given by the classifier, we will discard this label.

For active labeling, we generate pseudo labels only from negative labeled points at the first iteration. That is to say, if the pseudo label is not included in the negative label set, this point will not be taken as a pseudo label in this iteration. Notice that FSF and ACT are only utilized on the 3D network. The reason is the quality of the pseudo label produced by images is much worse. However, with the E-step, our approach can generate accurate pseudo labels, helping the two networks to get better optimized. With more accurate labels, the weak 2D network is hence boosted and could provide more useful information. The overall procedure is summarized in \cref{alg:main}.

\section{Experiments}
In this section, we perform experiments to present a comprehensive evaluation of our approach. 
\begin{table}[]
	\caption{Statistics of evaluation datasets.} \label{tab:statofdataset} 
	\resizebox{\linewidth}{!}{
		\begin{tabular}{@{}c|c|ccc@{}}
			\toprule
			&                  & nuScenes & SemanticKITTI & Waymo  \\ \midrule
			\multirow{3}{*}{Sensor}   & LiDARs           & 1                 & 1             & 1      \\
			& Images           & 6                 & 1             & 5      \\
			& Avg Points/Frame & 34K               & 120K          & 177K   \\ \midrule
			\multirow{2}{*}{Annot.} & Training Set     & 28,130            & 19,132        & 23,691 \\
			& Validation Set   & 6,019             & 4071         & 5,976  \\ \midrule
			\multirow{2}{*}{Subset} & Training Set     & 76.8\%             & 15.9\%         & 64.4\%  \\
			& Validation Set   & 76.8\%             & 16.3\%         & 64.3\%  \\ \bottomrule
		\end{tabular}
	}
\end{table}

\begin{table*}[]{
\caption{Quantitative results of different approaches on nuScenes validation set.}\label{tab:nusc}
\resizebox{\linewidth}{!}{
\begin{tabular}{@{}l|c|cccccccccccccccc|c@{}}
\toprule[1pt]
Methods                & Annot.               & \rotatebox{90}{barrier} & \rotatebox{90}{bicycle} & \rotatebox{90}{bus} & \rotatebox{90}{car} & \rotatebox{90}{construction} & \rotatebox{90}{motorcycle} & \rotatebox{90}{pedestrian} & \rotatebox{90}{traffic\_cone} & \rotatebox{90}{trailer} & \rotatebox{90}{truck} & \rotatebox{90}{driveable} & \rotatebox{90}{other\_flat} & \rotatebox{90}{sidewalk} & \rotatebox{90}{terrain} & \rotatebox{90}{manmade} & \rotatebox{90}{vegetation} & \rotatebox{90}{mIoU(\%)}      \\ \midrule
RangNet++('19)\cite{milioto2019rangenet++} & \multirow{8}{*}{100\%} & 66.0 & 21.3 & 77.2 & 80.9 & 30.2 & 66.8 & 69.6 & 52.1 & 54.2 & 72.3 & 94.1 & 66.6 & 63.5 & 70.1 & 83.1 & 79.8 & 65.5 \\
SPVCNN('20)\cite{tang2020searching}        &                        & 70.3 & 30.2 & 85.8 & 91.1 & 42.6 & 73.5 & 74.1 & 56.9 & 54.7 & 81.3 & 93.7 & 64.2 & 68.7 & 73.3 & 86.6 & 85.3 & 70.8 \\
PolarNet('20)\cite{zhang2020polarnet}      &                        & 74.7 & 28.2 & 85.3 & 90.9 & 35.1 & 77.5 & 71.3 & 58.8 & 57.4 & 76.1 & 96.5 & 71.1 & 74.7 & 74.0 & 87.3 & 85.7 & 71.0 \\
Salsanext('20)\cite{cortinhal2020salsanext} &                       & 74.8 & 34.1 & 85.9 & 88.4 & 42.2 & 72.4 & 72.2 & 63.1 & 61.3 & 76.5 & 96.0 & 70.8 & 71.2 & 71.5 & 86.7 & 84.4 & 72.2 \\
Cylinder3D('21)\cite{zhu2021cylindrical}    &                       & 74.5 & 43.1 & 87.4 & 85.9 & 45.1 & 80.2 & 79.7 & 65.3 & 61.5 & 80.6 & 96.5 & 71.2 & 74.9 & 75.3 & 87.7 & 87.1 & 74.8 \\
Cylinder3D('21)\cite{zhu2021cylindrical}$^\dagger$ &                & 76.4 & 40.3 & 91.3 & 93.8 & 51.3 & 78.0 & 78.9 & 64.9 & 62.1 & 84.4 & 96.8 & 71.6 & 76.4 & 75.4 & 90.5 & 87.4 & 76.1 \\
AMVNet('21)\cite{liong2020amvnet}$^\dagger$ &                       & 79.8 & 32.4 & 82.2 & 86.4 & 62.5 & 81.9 & 75.3 & 72.3 & 83.5 & 65.1 & 97.4 & 67.0 & 78.8 & 74.6 & 90.8 & 87.9 & 76.1 \\
RPVNet('21)\cite{xu2021rpvnet}$^\dagger$    &                       & 78.2 & 43.4 & 92.7 & 93.2 & 49.0 & 85.7 & 80.5 & 66.0 & 66.9 & 84.0 & 96.9 & 73.5 & 75.9 & 76.0 & 90.6 & 88.9 & 77.6 \\ \midrule
PMF('21)\cite{zhuang2021perception}   & \multirow{3}{*}{76.8\%} & 74.1 & 46.6 & 89.8 & 92.1 & 57.0 & 77.7 & 80.9 & 70.9 & 64.6 & 82.9 & 95.5 & 73.3 & 73.6 & 74.8 & 89.4 & 87.7 & 76.9 \\
PMF('21)\cite{zhuang2021perception}*  &  & 73.4 & 55.4 & 80.8 & 91.3 & 56.6 & 59.3 & 82.8 & 72.4 & 56.9 & 75.9 & 95.8 & 71.4 & 73.7 & 75.8 & 90.2 & 89.7 & 75.1 \\
\textbf{Ours Baseline}                &  & 75.4 & 57.6 & 92.0 & 92.3 & 62.5 & 85.8 & 83.6 & 72.6 & 64.1 & 82.6 & 95.5 & 73.5 & 72.8 & 74.6 & 90.2 & 88.9 & \textbf{79.0} \\ \midrule
SLidR('22)\cite{sautier2022image}* & \multirow{3}{*}{0.8\%} & 69.6 & 28.6 & 88.5 & 90.0 & 42.9 & 66.1 & 64.2 & 47.6 & 61.2 & 80.7 & 95.0 & 71.1 & 70.9 & 71.4 & 88.2 & 86.4 & 70.1 \\
Ours Baseline &                        & 73.8 & 50.4 & 84.5 & 90.9 & 53.0 & 80.9 & 76.7 & 61.7 & 48.9 & 74.1 & 94.8 & 71.3 & 71.6 & 73.6 & 89.1 & 87.7 & 73.9 \\
\textbf{Ours} &                        & 74.7 & 55.1 & 90.2 & 92.8 & 58.1 & 84.8 & 81.1 & 67.3 & 62.0 & 83.2 & 95.5 & 73.1 & 73.1 & 74.3 & 90.2 & 89.0 & \textbf{77.8} \\ \bottomrule[1pt]
\end{tabular}
}    
{\footnotesize * Results of these methods are reported by re-implementing their papers with officially released codes. $\dagger$ Results of these methods are calculated from the entire validation set as official codes are unavailable.}
}
\end{table*}

\subsection{Experimental Setting}
 \textbf{Datasets.} We empirically evaluate the performance of our method on the benchmark datasets including nuScenes \cite{caesar2020nuscenes}, SemanticKITTI \cite{behley2019semantickitti} and Waymo \cite{sun2020scalability}. The basic statistics of datasets are summarized in \cref{tab:statofdataset}. \textbf{nuScenes} contains 1,000 driving scenes collected from two different cities, Boston and Singapore, with different weather and light conditions. The scenes are of 20s each and are split into 28,130 training frames and 6,019 validation frames. Each frame contains a 32-beam LiDAR point cloud provided with pointwise annotations and six RGB images captured by six cameras from different views of LiDAR. 16 categories are used for segmentation. \textbf{SemanticKITTI} is a large-scale dataset based on KITTI Odometry Benchmark \cite{geiger2012we} captured in Germany. 43,000 scans with pointwise semantic annotations are provided, where 23,201 scans (sequence 00-10) are available for training (19,130 scans) and validation (4,071 scans). Unlike nuScenes, SemanticKITTI only provides the front-view images. 19 categories are used for segmentation. \textbf{Waymo} contains 2,030 segments of 20s each captured from Phoenix, San Francisco, and Mountain View, with diverse geographies and conditions. The segments are split into 800 training segments (23,691 frames in total) and 350 validation segments (5,976 frames in total). Each frame contains a high-resolution LiDAR point cloud captured by five LiDAR sensors (a mid-range LiDAR and four short-range LiDAR, we only use the points captured by the mid-range LiDAR) with pointwise annotations and five RGB images captured by five cameras from the front and side view of the car. 22 categories are used for segmentation. 
 
 
 \textbf{Cross-modal Data Processing.} To ensure every 3D points have its corresponding 2D pixels, following \cite{zhuang2021perception}, we remove all the points that are out of the cameras'view and therefore construct the corresponding subsets for training and evaluation. In fact, these subsets keep almost the same number of points for both training and validation sets. Therefore, they are able to reflect the entire data distribution. For example, as shown in \cref{tab:statofdataset}, for both training and validation sets, around 80\% of data in nuScenes is included in the subset. To make a fair comparison, all of the released models of the state-of-the-art methods are trained and evaluated under their original datasets and settings but we only collect the prediction of the points within our evaluation subset to calculate the final performance. 
 
\textbf{Evaluation Protocols.} Following the prior work \cite{qi2017pointnet}, we use the mean Intersection-over-Union (mIoU) as the main evaluation metric. We report the results under fully supervised and weakly supervised settings. If not specified, the baseline is trained with only the sparse label. Besides, we compare our method with random labeling in the ablation study, where we randomly select the same number of points for annotation.

\textbf{Iteration Protocols.} For active labeling, in the first iteration we only estimate pseudo labels in the points that are marked as negative labels. In the following iteration, we will resume the pseudo label in the previous iteration and generate new pseudo labels from unlabeled points.



\subsection{Implementation Details}
We use SPVCNN and SwiftNet as backbones of the LiDAR stream and camera stream, respectively. For nuScenes and Waymo, we initialize the parameters of SwiftNet with ResNet-18 \cite{paszke2019pytorch} pre-trained on ImageNet. For SemanticKITTI, we adapt SwiftNet to ResNet-34, and finetune it after pretraining on nuScenes, as SemanticKITTI only includes one camera which is insufficient for training. We adopt SGD with Nesterov \cite{nesterov1983method} as our optimizer for both 3D and 2D networks. 
For nuScenes, Waymo, and semanticKITTI, the initial learning rates for the 3D branch and fusion module are 0.24, 0.24, and 0.18 respectively. The initial learning rates for the 2D branch are 0.24, 0.004, and 0.0002. All decay to 0 with cosine policy \cite{loshchilov2016sgdr}. The batch size for all of the datasets is set to 12. As SPVCNN is based on 3D voxels, we set voxel size to 0.05m. The hyper-parameters $\alpha$ and $\delta$ in ACT are set to 0.5 and 0.1, respectively. The weight for $\mathcal{L}_{seg}$ and $\mathcal{L}_{asso}$ are set to 1.0 and 0.5. To implement the association learning, we use two projection heads to align the point cloud features and image embeddings, where the output dimension is set to 256. To prevent overfitting, we apply random rotation along the z-axis and random scaling on 3D point clouds.


\begin{table*}[]
\caption{Quantitative results of different approaches on semanticKITTI validation set}\label{tab:semKITTI}
\resizebox{\linewidth}{!}{
\begin{tabular}{@{}l|c|ccccccccccccccccccc|c@{}}
\toprule[1pt]
Methods         & Annot.               & \rotatebox{90}{car} & \rotatebox{90}{bicycle} & \rotatebox{90}{motorcycle} & \rotatebox{90}{truck} & \rotatebox{90}{other-vehicle} & \rotatebox{90}{person} & \rotatebox{90}{bicyclist} & \rotatebox{90}{motorcyclist} & \rotatebox{90}{road} & \rotatebox{90}{parking} & \rotatebox{90}{sidewalk} & \rotatebox{90}{other-ground} & \rotatebox{90}{building} & \rotatebox{90}{fence} & \rotatebox{90}{vegetation} & \rotatebox{90}{trunk} & \rotatebox{90}{terrain} & \rotatebox{90}{pole} & \rotatebox{90}{traffic-sign} & \rotatebox{90}{mIoU(\%)} \\ \midrule
RangNet++('19)\cite{milioto2019rangenet++} & \multirow{5}{*}{100\%} & 89.4 & 26.5 & 48.4 & 33.9 & 26.7 & 54.8 & 69.4 & 0.0 & 92.9 & 37.0 & 69.9 & 0.0 & 83.4 & 51.0 & 83.3 & 54.0 & 68.1 & 49.8 & 34.0 & 51.2 \\
RandLANet('20)\cite{hu2020randla} &                        & 92.0 & 8.0 & 12.8 & 74.8 & 46.7 & 52.3 & 46.0 & 0.0 & 93.4 & 32.7 & 73.4 & 0.1 & 84.0 & 43.5 & 83.7 & 57.3 & 73.1 & 48.0 & 27.3 & 50.0 \\
SPVCNN('20)\cite{tang2020searching} &                        & 96.0 & 41.2 & 62.0 & 32.9 & 53.4 & 67.3 & 82.0 & 0.0 & 91.9 & 31.0 & 76.0 & 1.1 & 87.5 & 47.8 & 84.8 & 62.0 & 65.3 & 64.1 & 49.7 & 64.5 \\
Salsanext('20)\cite{cortinhal2020salsanext} &                   & 90.5 & 44.6 & 49.6 & 86.3 & 54.6 & 74.0 & 81.4 & 0.0 & 93.4 & 40.6 & 69.1 & 0.0 & 84.6 & 53.0 & 83.6 & 64.3 & 64.2 & 54.4 & 39.8 & 59.4 \\
Cylinder3D('21)\cite{zhu2021cylindrical} &                        & 96.8 & 61.1 & 82.7 & 64.7 & 74.5 & 82.8 & 94.4 & 0.0 & 95.4 & 46.9 & 79.0 & 2.1 & 88.3 & 53.1 & 86.9 & 73.2 & 69.0 & 63.5 & 41.2 & \textbf{65.9} \\ \midrule
PMF('21)\cite{zhuang2021perception}* & \multirow{4}{*}{15.9\%} & 95.4 & 53.1 & 58.6 & 29.1 & 62.3 & 80.8 & 88.3 & 0.8 & 96.0 & 40.0 & 80.3 & 0.4 & 88.6 & 59.2 & 87.5 & 72.8 & 71.5 & 64.7 & 44.6 & 61.8 \\
PMF('21)\cite{zhuang2021perception} & & 95.4 & 47.8 & 62.9 & 68.4 & 75.2 & 78.9 & 71.6 & 0.0 & 96.4 & 43.5 & 80.5 & 0.1 & 88.7 & 60.1 & 88.6 & 72.7 & 75.3 & 65.5 & 43.0 & 63.9 \\
SPVCNN('20)\cite{tang2020searching} & & 95.8 & 40.3 & 61.8 & 75.6 & 59.8 & 75.3 & 87.9 & 0.0 & 94.8 & 37.3 & 77.1 & 0.1 & 87.1 & 55.4 & 87.9 & 65.2 & 73.2 & 63.2 & 40.5 & 62.0 \\
\textbf{Ours}                       & & 97.0 & 43.0 & 62.1 & 76.7 & 71.5 & 84.7 & 89.9 & 0.0 & 95.6 & 38.5 & 79.3 & 0.2 & 89.3 & 59.4 & 88.4 & 73.6 & 73.3 & 67.2 & 43.8 & \textbf{64.9} \\ \midrule
Scribble('22)\cite{unal2022scribble} & 8\% & 91.2 & 42.1 & 64.8 & 80.7 & 65.6 & 81.4 & 88.5 & 0.0 & 90.0 & 32.1 & 70.6 & 3.1 & 88.2 & 53.0 & 86.3 & 68.6 & 69.5 & 61.5 & 43.3 & 61.9 \\
ReDAL('21)\cite{wu2021redal}$^\dagger$ & 5\% & 95.4 & 29.6 & 58.6 & 63.4 & 49.8 & 63.4 & 84.1 & 0.5 & 91.5 & 39.3 & 78.4 & 1.2 & 89.3 & 54.4 & 87.4 & 62.0 & 74.1 & 63.5 & 49.7 & 59.8 \\
HybridCR('22)\cite{li2022hybridcr}$^\dagger$ & 1\%   & - & - & - & - & - & - & - & - & - & - & - & - & - & - & - & - & - & - & - & 51.9 \\
SQN('21)\cite{hu2021sqn}$^\dagger$ & 0.1\% & 92.1 & 39.3 & 30.1 & 36.7 & 26.0 & 36.4 & 25.3 & 7.2 & 90.5 & 56.8 & 72.9 & 19.1 & 84.8 & 53.3 & 80.8 & 59.1 & 67.0 & 44.5 & 44.0 & 50.8 \\
SLidR('22)\cite{sautier2022image}* & 0.08\% & 94.1 & 5.8 & 26.8 & 70.9 & 57.1 & 62.0 & 81.6 & 0.0 & 94.7 & 34.6 & 75.1 & 0.2 & 85.3 & 40.3 & 83.1 & 55.9 & 66.1 & 49.0 & 25.7 & 53.1 \\
Ours Baseline & 0.08\% & 95.0 & 45.3 & 62.7 & 60.9 & 62.7 & 68.4 & 74.9 & 0.0 & 94.6 & 37.1 & 77.4 & 1.6 & 87.7 & 52.2 & 86.4 & 67.6 & 70.0 & 59.0 & 41.5 & 60.2 \\
\textbf{Ours} & 0.08\% & 95.6 & 48.7 & 63.5 & 75.6 & 67.3 & 75.3 & 79.8 & 0.0 & 95.0 & 38.9 & 78.2 & 2.0 & 89.9 & 62.1 & 88.2 & 71.6 & 71.9 & 62.7 & 45.1 & \textbf{63.7} \\ \bottomrule[1pt]
\end{tabular}
}
{\footnotesize * Results of these methods are reported by re-implementing their papers with officially released codes. $\dagger$ Results of these methods are calculated from the entire validation set as official codes are unavailable.}
\end{table*}

\begin{table}[]
\centering
\caption{Quantitative results of different approaches on waymo validation set.}
 \resizebox{0.75\linewidth}{!}{
\begin{tabular}{@{}l|c|c|c@{}}
\toprule
Methods                & Modality & Annot.                 & mIoU(\%)       \\ \midrule
SPVCNN('20) \cite{tang2020searching} & \multirow{2}{*}{LiDAR} & \multirow{2}{*}{100\%}  & 65.5 \\
Cylinder3D('21) \cite{zhu2021cylindrical} &                &                        & 62.6 \\ \midrule
PMF('21) \cite{zhuang2021perception} & \multirow{4}{*}{\begin{tabular}[c]{@{}c@{}}LiDAR\\ +\\ Camera\end{tabular}} & \multirow{2}{*}{64.4\%} & 58.2 \\
\textbf{Ours Baseline} &                                    &                         & \textbf{67.0} \\ \cmidrule(r){1-1} \cmidrule(l){3-4}
Ours Baseline          &                                    & \multirow{2}{*}{0.3\%}  & 63.5 \\
\textbf{Ours}          &                                    &                         & \textbf{65.7} \\ \bottomrule
\end{tabular}
 }
\label{tab:waymo}
\end{table}

\subsection{Results on nuScenes}
We report the quantitative results on nuScenes in \cref{tab:nusc}. It is observed that even with 0.8\% of sparse labeled points, our method beats almost all the recent fully supervised methods. Specifically, under the fully supervised setting, our baseline with image fusion outperforms the state-of-the-art LiDAR-only methods like RPVNet \cite{xu2021rpvnet} by \textbf{1.4\%} in mIoU. It indicates information contained in images can well complement and assist LiDAR data. Our method also beats the prior cross-modal method PMF \cite{zhuang2021perception} by \textbf{2.1\%} in mIoU, demonstrating our baseline can well combine the information between 3D voxels and images. 
Under the 0.8\% label setting, our baseline achieves 73.9\%@mIoU. But with the help of our active labeling and cross-modal self-training framework, our baseline receives \textbf{3.9\%} performance gain and achieves 77.8\% in mIoU which is highly comparable with the state-of-the-art supervised methods. 

To evaluate the weakly supervised performance of our method, we also compare our method to the most related work SLidR \cite{sautier2022image}, which also utilizes the idea of the superpixel. This is done in a label-agnostic contrastive learning way. The main difference is 1) SLidR freezes most of the weight of the 2D branch during training while these weights are updated in our method, which allows us to train the whole network jointly; 2) in contrast to contrastive learning, our EM self-training strategy could consistently improve the performance. Therefore, our learned image feature is more compatible with the 3D branch. As the label setting of SLidR is different from ours, hence, to make a fair comparison, we adapt SLidR by training it with the same percentage of randomly sampled labels. The class distribution of the label is guaranteed by proportionally selecting the label from each class. As shown in \cref{tab:nusc}, with the same amount of annotation, our method surpasses SLidR by 7.7@mIoU, demonstrating the effectiveness of our cross-modal self-training framework. 

\subsection{Results on SemanticKITTI}
As shown in \cref{tab:semKITTI}, we report our quantitative results on SemanticKITTI. Note that this dataset only provides the front-view images, so we have to only use the front-view point cloud for training and testing, following prior cross-modality work PMF  \cite{zhuang2021perception}. Even in this situation, our model still increases to 64.9@mIoU 1\% higher than PMF \cite{zhuang2021perception} and only 1\% lower than the state-of-the-art lidar-only method Cylinder3D \cite{zhu2021cylindrical} which is trained on the full training set. To make a fair comparison, we pretrain PMF on the nuScenes training set with full label, and then finetune the network on SemanticKITTI, as same as our method. However, we only obtain 61.8\% in terms of mIoU, as shown in \cref{tab:semKITTI} PMF*.

Another observation is that many existing weakly supervised methods often adopt RandLANet as the baseline, and its performance is somewhat weak. 
Thanks to our strong fusion baseline, we outperform them by a large margin. We also compare our method with recently proposed voxel-based backbones. For example, Scribble \cite{unal2022scribble} adopt scribbles for LiDAR point cloud which is popular in 2D semantic segmentation and achieves 61.9@mIoU with 8\% annotated points. 
ReDAL \cite{wu2021redal} employs active labeling, reaching 59.8@mIoU with 5\% annotations. Our method surpasses all these mentioned methods by achieving 63.7@mIoU with the least annotation of 0.08\%. Compared with our baseline method only trained with sparse labels, our full method receives a 3.5@mIoU performance gain. These experiment results also support our claim that combining LiDAR and images can bring benefits, especially in the weakly supervised setting.


\begin{figure*}[p]
\begin{adjustbox}{addcode={
\begin{minipage}{\width}}{
\caption{Visualization of the active labeling results. The left and right parts of the figure show two individual examples. Fig. (a) shows the pre-segment result where points in the same color belong to the same cluster. Fig. (b) shows the sparse label based on the pre-segment result where points in black are unlabeled. Fig. (c) shows the propagate label. Fig. (d) shows the ground truth label. Fig. (e) shows the points covered by at least one of the three types of labels.}
\label{fig:vis_al}
\end{minipage}},rotate=90,center}
\includegraphics[width=0.99\textheight]{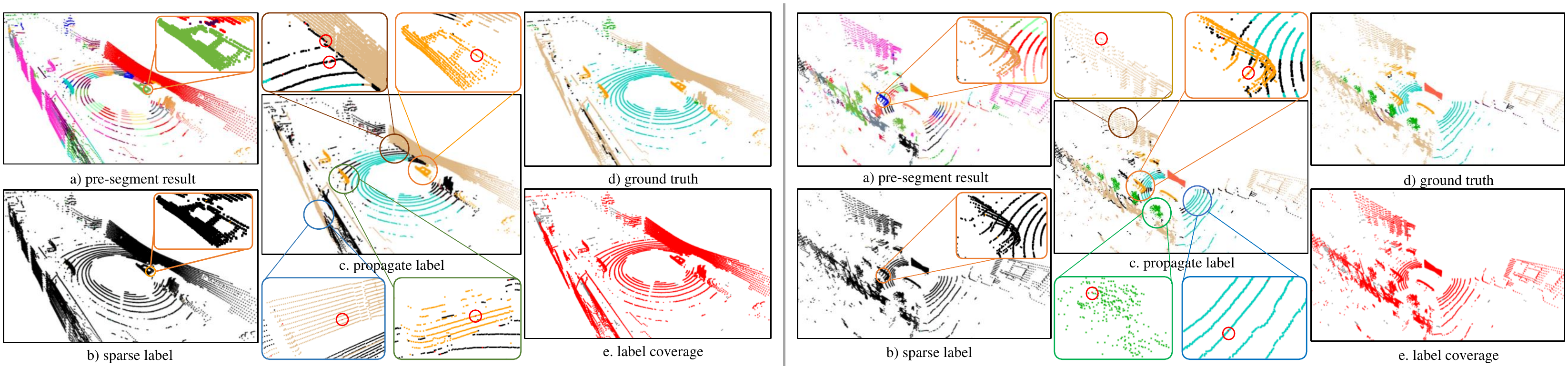}
\end{adjustbox}
\end{figure*}

\begin{figure}[t]
  \centering
  \includegraphics[width=0.99\linewidth]{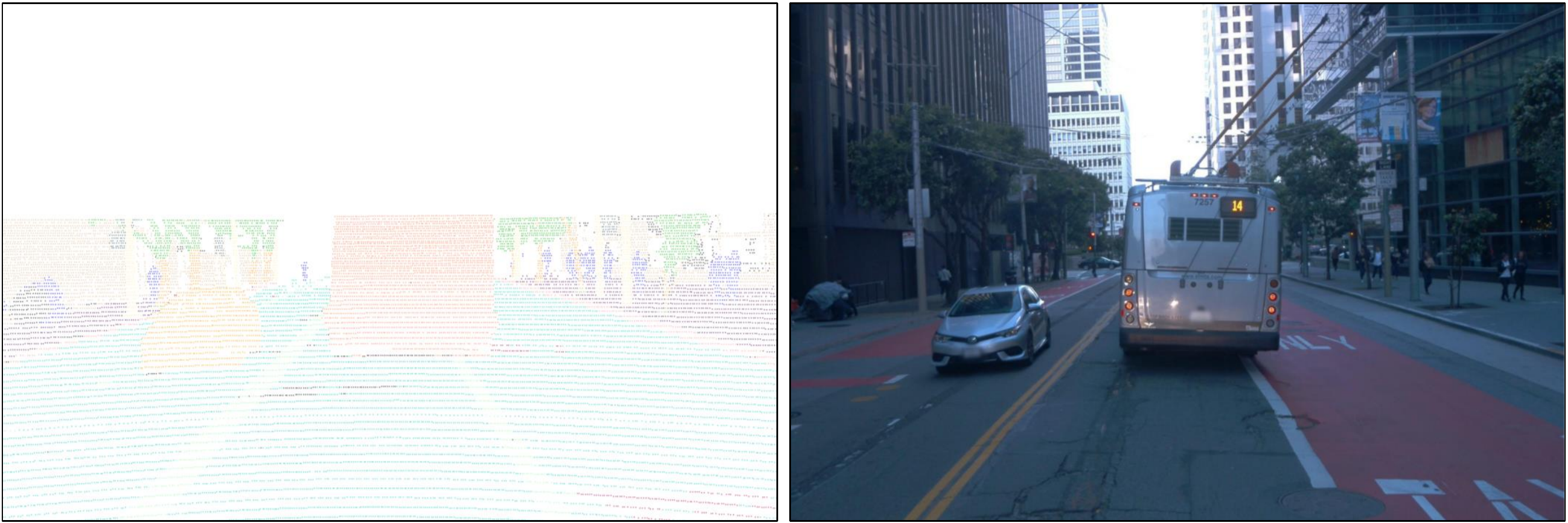}
  \caption{
  The left image shows the projected point cloud from Waymo that will be fed into the 3D branch of PMF. The right image shows the corresponding image from the camera. As shown above, most of the pixels are empty which causes the sub-optimal solution.}
  \label{fig:pmfempty}
\end{figure}
\begin{figure}[t]
  \centering
  \includegraphics[width=0.99\linewidth]{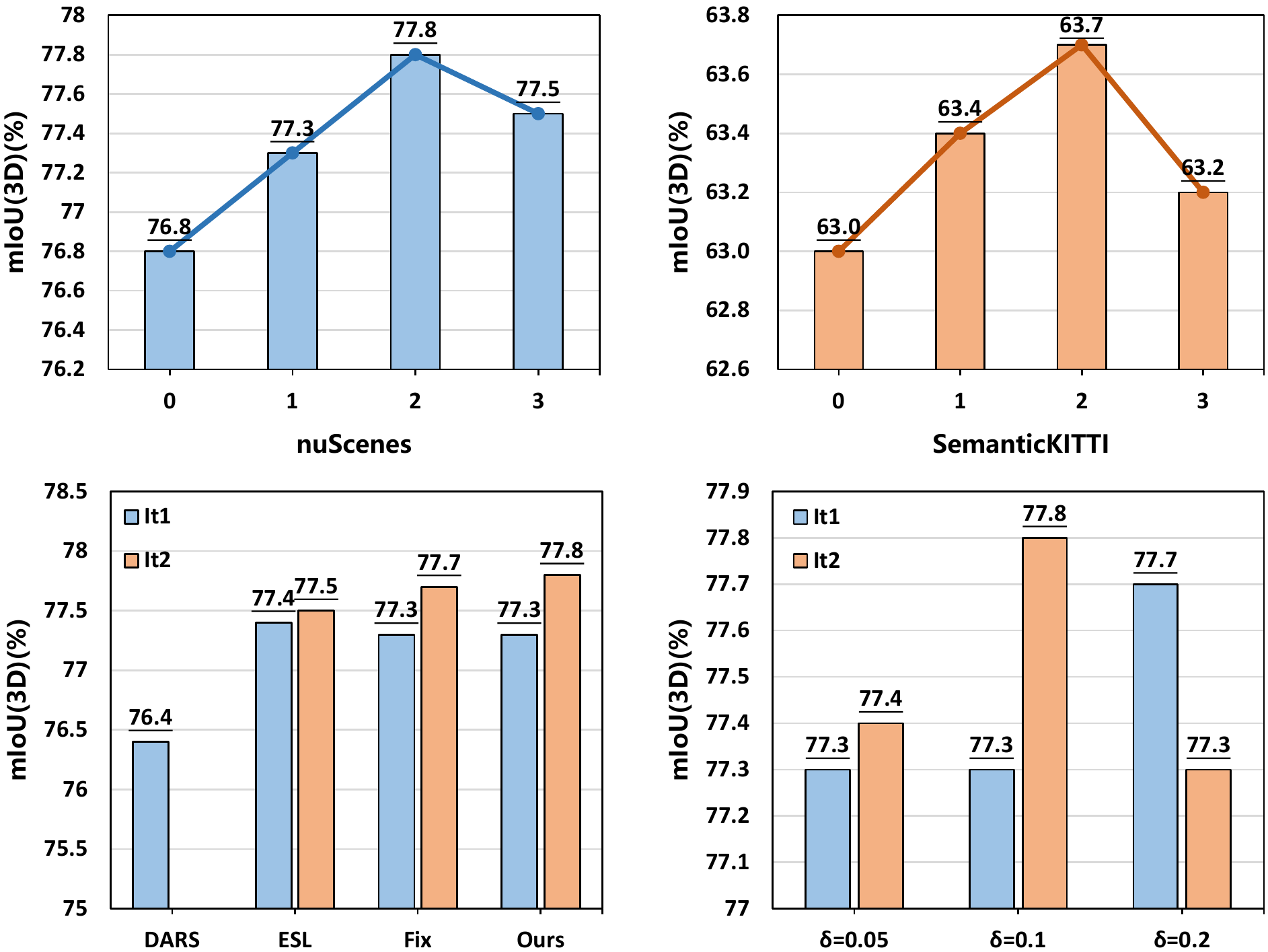}
  \caption{The top two figures show the iteration training experiments on nuScenes and SemanticKITTI. The two figures below show the ablation experiments for E-step. The left figure shows the comparison between other pseudo label methods and our self-rectification module. Fix for fixed confidence threshold, ESL \cite{saporta2020esl} for Entropy-guided method, DARS \cite{he2021re} for distribution align based method. The right figure shows the results when varying confidence tolerance $\delta$.}
  \label{fig:iteration}
\end{figure}

\subsection{Results on Waymo}

As shown in \cref{tab:waymo}, we further report our quantitative results on Waymo. As it is a recently released dataset, the official results of most methods are unavailable. Therefore, we report the results by re-implementing their papers with officially released codes. Here are the key observations: 1) our baseline achieves the highest performance in the fully supervised manner. Similar to nuScenes, our method outperforms SPVCNN by 1.5@mIoU by using the complementary information from images.  
2) Our method also outperforms the cross-modality method PMF by a large margin. We find PMF relies on projecting sparse LiDAR point cloud to dense perspective view images, but most of the pixels are empty on Waymo, which leads to a sub-optimal solution shown in Fig. \ref{fig:pmfempty}.
3) Under the weakly supervised setting, we obtain 2.2@mIOU gain due to the self-training framework.


\subsection{Ablation Study}
In this subsection, we conduct ablation studies to further verify the effectiveness of each component of our method. For a fair comparison, we follow the same experimental settings with 0.8\% active labeling on nuScenes. The main results are shown in \cref{tab:ablation}, \cref{tab:cml} and \cref{fig:iteration}.

\textbf{Effectiveness of Active Labeling.} As shown in \cref{tab:ablation}, in line \#1, we report the performance (73.4@mIoU) using random labeling, where we proportionally sample the points from each category to keep the class distribution. Note that the number of labels in random labeling is the same as that in active labeling. Here are key observations: 1) Training with only sparse labels already outperforms the random labeling by 0.5@mIoU. 2) Negative label provides denser supervision thus improving the performance to 74.9@mIoU(\#3 Row). Propagate label provides more accurate supervision and the performance reach 76.2@mIoU(\#4 Row). In summary, compared with random labeling, our active labeling absorbs valuable human supervision based on the geometric prior of the LiDAR point cloud and enables us to significantly reduce the labeling efforts. 

As shown in \cref{fig:vis_al}, we randomly sample two point clouds from nuScenes to show how the point cloud is labeled with our method. \cref{fig:vis_al}. a, b, c, d denote the whole active labeling results. For example, in c, we show the sparse labeled points in red and emphasize them with a red circle. The colored points are annotated with propagated labels. As we can see, instead of carefully labeling the entire car, now only a single click can do the same job, which can significantly save the time and annotation budget.

\begin{figure*}[t]
\centering
\includegraphics[width=0.99\linewidth]{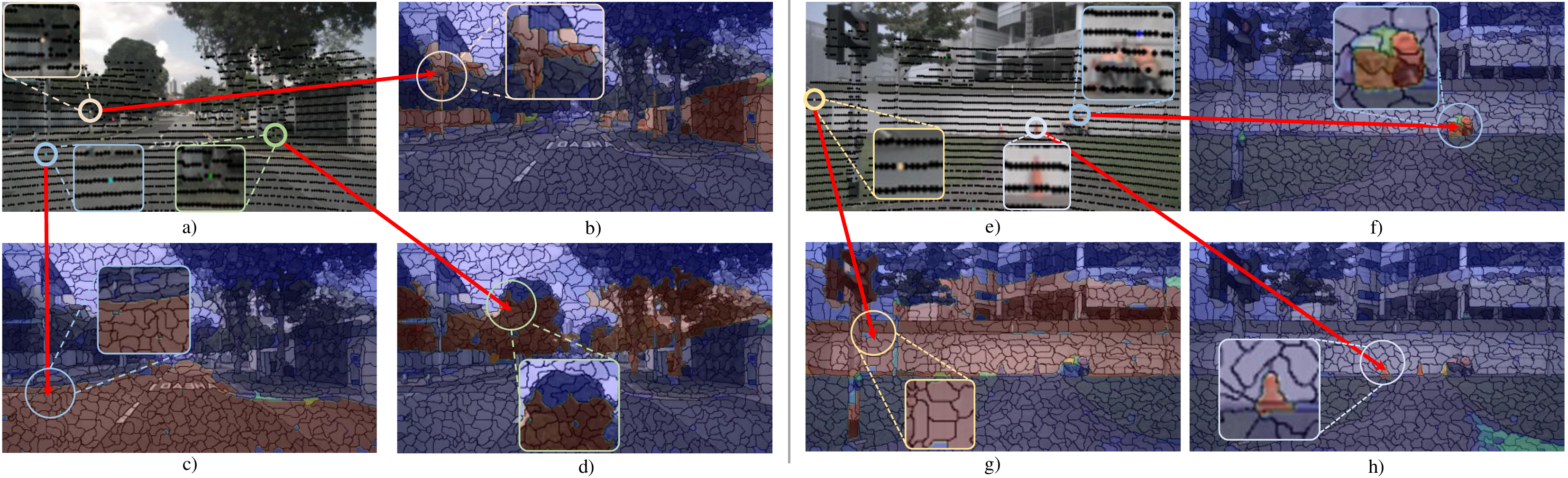}
\caption{Visualization of the association results. We compute the feature similarity from a randomly selected 3D point to all 2D superpixels. Blue to red represents the similarity from low to high. a) illustrates the selected 3D points (blue circle indicates a point on the road, orange on the wall, and green on the tree). b) to d) present the feature similarity of every superpixel to the point on the wall, road, and tree, respectively. e) to f) shows more association results. Note that even for small categories like the pedestrian (f) and the traffic cone (h), our method is able to successfully build the correlation between the two modalities.}
\label{fig:vis_asso}
\end{figure*}

\begin{table}[]
\caption{Comparison between different cross-modal learning methods. +xMUDA and +SpCML stand for adding their cross-modal learning loss based on our baseline. +PMF stands for adding its perception-aware loss based on our baseline.}\label{tab:cml}
\resizebox{\linewidth}{!}{
\begin{tabular}{@{}c|ccccc@{}}
\toprule
         & baseline & +xMUDA \cite{jaritz2020xmuda} & +SpCML \cite{peng2021sparse} & +PMF \cite{zhuang2021perception} & +CMAsL \\ \midrule
mIoU(3D) & 76.2     & 76.0       & 76.5   & 76.1     & 76.8   \\ \midrule
mIoU(2D) & 61.2     & 61.6       & 61.3   & 60.9     & 61.1   \\ \bottomrule
\end{tabular}
}
\end{table}

\textbf{Effectiveness of CMAsL.} As shown in \cref{tab:ablation} line \#5, with the help of CMAsL, 0.6@mIoU performance gain can be observed with our strong baseline. To further prove our method can better transfer the complementary knowledge from image to point cloud, we compare our method with related cross-modal learning methods including xMUDA \cite{jaritz2020xmuda}, DsCML \cite{peng2021sparse}, PMF \cite{zhuang2021perception} the results are reported in \cref{tab:cml}. 
xMUDA \cite{jaritz2020xmuda} employs mutual learning directly on the prediction from the 2D and 3D branches. 
SpCML is adapted from DsCML \cite{peng2021sparse}, where it pools the feature from the corresponding superpixel if the corresponding points only come from one category. Then we employ mutual learning on the predictions from 3D points and 2D superpixels. PMF \cite{zhuang2021perception} estimates entropy based on the prediction from both branches and conducts prediction aligning from high entropy modal to low entropy one. As shown in \cref{tab:cml}, xMUDA and PMF fail to bring any benefit for the 3D branch, which shows the naive combination of the two branches would degrade the performance because of the imbalanced modality capability. Though SpCML also improves the performance with the help of superpixel pooling, the improvement is slight. Notice that in our design, CMAsL mainly uses prior knowledge from images and hence brings no benefits for the 2D branch.

\textbf{Effectiveness of E-step.}  In the E-Step, we use the self-training framework to generate pseudo labels, and then use these labels to fine-tune the network. From the last line in \cref{tab:ablation}, iteration training with ACT and FSF module improve the performance by 1.0@mIoU. Furthermore, we compare our method with other representative pseudo label filtering methods. The results are shown in \cref{fig:iteration}. ESL \cite{saporta2020esl} filters out the points with their entropy larger than a given threshold. The threshold is set to be the median of the entropy of the points from each category. DARS \cite{he2021re} proves to be very effective in semi-supervised 2D semantic segmentation. It sets the confidence threshold for each category to select a group of pseudo labels that keep the same distribution as the labeled dataset. In our experiments, we also compare a fixed confidence threshold method in our ACT module denoted as FIX.
Along with all of the methods, our proposed method performs slightly better. Steady improvement can be found in every iteration. It appears that ESL can hardly bring any benefit in the 2nd iteration, while DARS even performs slightly worse as the category imbalance issue on 3D datasets is even more significant than that on 2D datasets. It causes DARS to be very aggressive in minor classes which introduce many noise labels. 

\begin{table}[]
\caption{Ablation study on nuScenes validation set. \textbf{sp} denotes using sparse label. \textbf{pp} denotes using propagate label. \textbf{neg} denotes using negative label. In the first line, we report the result of randomly labeling the same percentage of points as the sparse label as the baseline.}\label{tab:ablation}
\resizebox{\linewidth}{!}{
\begin{tabular}{@{}c|ccccc|c|c@{}}
\toprule
    & \multicolumn{5}{c|}{Components}                                & \multirow{2}{*}{mIoU(3D)} & \multirow{2}{*}{mIoU(2D)} \\
    & sp         & pp         & neg        & CMASL      & EStep      &                           &                           \\ \midrule
\#1 &            &            &            &            &            & 73.4                      & 59.5                      \\
\#2 & \checkmark &            &            &            &            & 73.9                      & 54.8                      \\
\#3 & \checkmark &            & \checkmark &            &            & 74.9                      & 56.8                      \\
\#4 & \checkmark & \checkmark & \checkmark &            &            & 76.2                      & 61.2                      \\
\#5 & \checkmark & \checkmark & \checkmark & \checkmark &            & 76.8                      & 61.1                      \\
\#6 & \checkmark & \checkmark & \checkmark & \checkmark & \checkmark & 77.8                      & 62.3                      \\ \bottomrule
\end{tabular}
}
\end{table}

\begin{figure*}[t]
  \centering
  \includegraphics[width=0.99\linewidth]{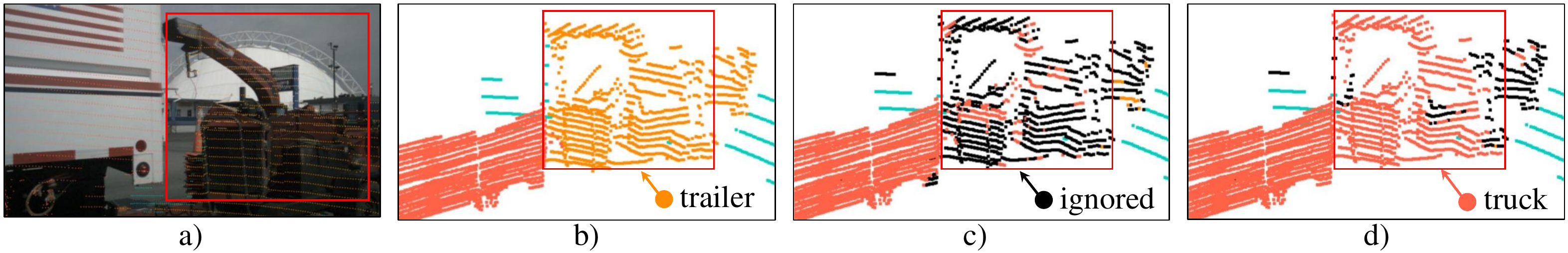}
  \caption{Visualization of the pseudo label estimation results with and without self-rectification. a) and b) present the ground truth labels in the image view and point cloud view. c) presents the pseudo labels with our self-rectification module and d) is the pseudo labels with only a fixed confidence threshold filter.}
  \label{fig:vis_fsf}
\end{figure*}

\begin{figure*}[p]
\begin{adjustbox}{addcode={
\begin{minipage}{\width}}{
\caption{Qualitative comparison on nuScenes (top), SemanticKITTI (mid), and Waymo (bottom). The red circles indicate the difference between the results of Our method and other competitors in weakly supervised settings (0.8\% annotation for nuScenes, 0.08\% annotation for SemanticKITTI, and 0.3\% for Waymo).}
\label{fig:vis_cmp_slidr}
\end{minipage}},rotate=90,center}
\includegraphics[width=0.99\textheight]{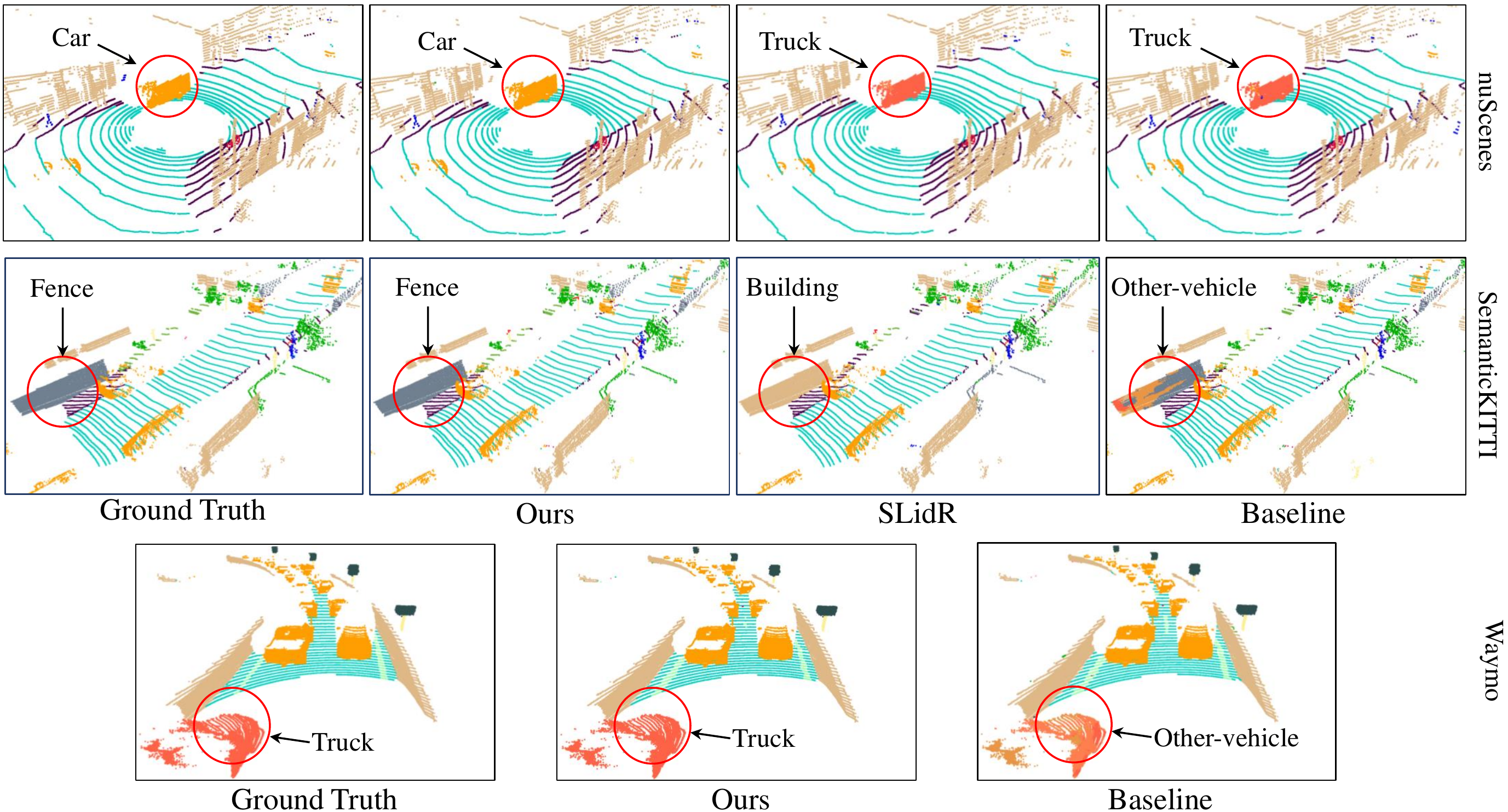}
\end{adjustbox}
\end{figure*}

\subsection{Visualization}
In this section, visualization results are provided to intuitively demonstrate the effectiveness of our approach.



\textbf{Cross-modal Association.} As shown in \cref{fig:vis_asso}, we compute the feature similarity from a randomly selected labeled 3D point to the 2D superpixels (center point) using the association projection heads. The results are shown in a heat map style. The red areas in the pictures represent high similarity and the blue areas are on the contrary. As we can see, the areas with high similarity in two kinds of data contain the same object, \textit{e.g.,} the road in \cref{fig:vis_asso}. a) and c). That indicates our association framework can strengthen the connection between two kinds of data. 

\textbf{Pseudo Label Self-rectification.} \cref{fig:vis_fsf} visualizes the effects of our pseudo label self-rectification module. \cref{fig:vis_fsf}. a) and b) show the ground truth label in the image view and point cloud view.
 \cref{fig:vis_fsf}. c) and d) show the estimated pseudo labels with our pseudo label self-rectification module and with a fixed confidence filter. The black points represent filtered points. As we can see in the red box in \cref{fig:vis_fsf}. d), large plenty of points are mislabeled and can not be filtered out by the fixed confidence filter, while our methods can keep the correct estimation in \cref{fig:vis_fsf}. c). This result demonstrates that our pseudo label self-rectification module is able to reduce the error rate of the pseudo labels.

\textbf{Qualitative results comparison.} Finally, we show the segment results of our method, SLidR, and our baseline in the validation set of nuScenes (top) and SemanticKITTI (mid), and show the segment results of our method and our baseline in the validation set of Waymo (bottom). All in weakly supervised settings in \cref{fig:vis_cmp_slidr}. Both SLidR and our baseline perform much worse on distinguishing similar categories (\textit{e.g.} car and truck, fence and building), while our method can avoid such misclassifications due to the better combination of the 2D and 3D data.

\section{Conclusion}
In this paper, we investigate a new cross-modal weakly supervised setting for 3D segmentation, and propose a
cross-modal baseline. With this baseline, we design a self-training solution with several critical improvements. The final framework outperforms all other state-of-the-art methods by a large margin on three popular datasets. We believe the new setting has great potential to be further explored for 3D segmentation. It is in the prospect that more efficient self-training methods can be designed with better representation power and less computational cost.

~\\
{\small\noindent \textbf{Availability of data and materials} The authors confirm that the datasets analysed as part of this research are freely available under the terms and conditions detailed in the license agreement enclosed in the data repository. nuScenes is available in \href{https://www.nuscenes.org}{https://www.nuscenes.org}. SemanticKITTI is available in \href{http://semantic-kitti.org}{http://semantic-kitti.org}. Waymo is available in \href{https://waymo.com/open}{https://waymo.com/open}.}

\section{Appendix}
The detailed deduction for our EM-based self-training framework:

To the optimize target described in \cref{eq:src_target}, we introduce an auxiliary probability distribution $q(\cdot)$ related to $\bar{\boldsymbol{y}}$, as it's difficult to optimize this function directly. Hence the objective is converted to:
\begin{equation}
    \begin{aligned}        \boldsymbol{\theta}^*&=\mathop{argmax}\limits_{\boldsymbol{\theta}}\sum\limits_{i=1}^N\log\sum\limits_{c=1}^C{q(\bar{y}^{(c)})\frac{p(\boldsymbol{x}^{(i)},\bar{y}^{(c)}|\boldsymbol{\theta})}{q(\bar{y}^{(c)})}} \\
                         &\geq\mathop{argmax}\limits_{\boldsymbol{\theta}  }\sum\limits_{i=1}^N\sum\limits_{c=1}^C{q(\bar{y}^{(c)})\log{\frac{p(\boldsymbol{x}^{(i)},\bar{y}^{(c)}|\boldsymbol{\theta})}{q(\bar{y}^{(c)})}}}.
    \end{aligned}
\end{equation}
The above is obtained by Jensen's inequality, and the equation holds if and only if $p(\boldsymbol{x}^{(i)},\bar{y}^{(c)}|\boldsymbol{\theta})/q(\bar{y}^{(c)})$ is a constant $K$ when $\sum_{c=1}^C{q(\bar{y}^{(c)})}=1$.
Therefore, we have:
\begin{equation}
 q(\bar{y}^{(c)})= \frac{p(\boldsymbol{x}^{(i)},\bar{y}^{(c)}|\boldsymbol{\theta})}{p(\boldsymbol{x}^{(i)})} =p(\bar{y}^{(c)}|\boldsymbol{x}^{(i)},\boldsymbol{\theta}).
\end{equation}
Finally, the optimization objective is redirected to:
\begin{equation}
    \boldsymbol{\theta}^*=\mathop{argmax}\limits_{\boldsymbol{\theta}}\sum\limits_{i=1}^N\sum\limits_{c=1}^C{p(\bar{y}^{(c)}|\boldsymbol{x}^{(i)},\boldsymbol{\theta})\log{p(\boldsymbol{x}^{(i)},\bar{y}^{(c)}|\boldsymbol{\theta})}}.
\end{equation}

\bibliographystyle{spmpsci}      
\bibliography{egbib}   
\end{sloppypar}
\end{document}